# Applying a Random Projection Algorithm to Optimize Machine Learning Model for Breast Lesion Classification

Morteza Heidari[1], Sivaramakrishnan Lakshmivarahan[2], Seyedehnafiseh Mirniaharikandehei[1], Gopichandh Danala[1], Sai Kiran R. Maryada[2], Hong Liu[1], Bin Zheng[1]

*Abstract*— Machine learning is widely used in developing computer-aided diagnosis (CAD) schemes of medical images. However, CAD usually computes large number of image features from the targeted regions, which creates a challenge of how to identify a small and optimal feature vector to build robust machine learning models. In this study, we investigate feasibility of applying a random projection algorithm to build an optimal feature vector from the initially CAD-generated large feature pool and improve performance of machine learning model. We assemble a retrospective dataset involving 1,487 cases of mammograms in which 644 cases have confirmed malignant mass lesions and 843 have benign lesions. A CAD scheme is first applied to segment mass regions and initially compute 181 features. Then, support vector machine (SVM) models embedded with several feature dimensionality reduction methods are built to predict likelihood of lesions being malignant. All SVM models are trained and tested using a leave-one-case-out cross-validation method. SVM generates a likelihood score of each segmented mass region depicting on one-view mammogram. By fusion of two scores of the same mass depicting on two-view mammograms, a case-based likelihood score is also evaluated. Comparing with the principle component analyses, nonnegative matrix factorization, and Chi-squared methods, SVM embedded with the random projection algorithm yielded a significantly higher case-based lesion classification performance with the area under ROC curve of 0.84±0.01 (p<0.02). The study demonstrates that the random project algorithm is a promising method to generate optimal feature vectors to help improve performance of machine learning models of medical images.

*Index Terms*— breast cancer diagnosis, computer-aided diagnosis (CAD) of mammograms, feature dimensionality reduction, lesion classification, random projection algorithm, support vector machine (SVM).

## I. INTRODUCTION

DEVELOPING computer-aided detection and diagnosis (CAD) schemes of medical images have been attracting broad research interest in order to detect suspicious diseased regions or patterns, classify between malignant and benign lesions, quantify disease severity, and predict disease prognosis or monitor treatment efficacy. Some of CAD schemes have been used as "a second reader" or quantitative image feature or marker assessment tools in current clinical practice to assist physicians (i.e., radiologists) reading and interpreting medical image, which aims to improve accuracy and/or efficiency of reading medical images, as well as reduce inter-reader variability [1]. Despite of extensive research effort and progress made in the CAD field, there are many remaining challenges or hurdles in developing CAD schemes for clinical applications [2]. For example, in developing CAD schemes, machine learning plays a critical role, which use image features to train classification models to predict the likelihood of the analyzed regions depicting or patterns representing diseases. However, due to the great heterogeneity of disease patterns and the limited size of training image datasets, how to identify a small and optimal image feature vector to build the highly performed and robust machine learning models remains a difficult task.

In current CAD schemes, after image preprocessing to reduce image noise, detecting and segmenting suspicious disease related regions of interest (ROIs), CAD schemes compute many image features from the entire image region or the segmented ROIs. For example, based on the recently developed radiomics concept and methods, more than 1,000 image features can be computed, which mostly represent texture patterns of the segmented ROIs in variety of scanning directions [3, 4]. However, due to the limited size of the training image datasets, such large number of image features can often drive to overfitting rather than learning the actual basis of a decision in building robust machine learning models. Thus, it is important to build an optimal feature vector from the initially large feature pool in which the generated features should not be redundant or highly correlated [5]. Then, machine learning models can be better trained with the enhanced performance and robustness. In general, if the feature dimensionality reduction happens with choosing the most effective image features from the initial feature pool, it is known as feature selection (i.e., using sequential forward

This work is supported in part by Grant No. R01-CA197150 from the National Cancer Institute, National Institutes of Health, USA. The authors would also like to acknowledge the support received from the Peggy and Charles Stephenson Cancer Center, University of Oklahoma, USA.

*M. Heidari, S. Mirniaharikandehei, and G. Danala are with the School of Electrical and Computer Engineering, University of Oklahoma, Norman, OK 73019, USA (e-mail: morteza.heidari@ou.edu, Snmirnia@ou.edu, and danala@ou.edu).

S. Lakshmivarahan, and S. Maryada are with School of Computer Science, University of Oklahoma, Norman, OK 73019, USA (e-mail: varahan@ou.edu and sai.maryada@ou.edu).

H. Liu and B. Zheng are with School of Electrical and Computer Engineering, University of Oklahoma, Norman, OK 73019, USA (e-mails: liu@ou.edu and bin.zheng-1@ou.edu).

floating selection (SFFS) [6]). On the other hand, if the dimensionality reduction comes from reanalyzing the initial set of features to produce a new set of orthogonal features, it is known as feature regeneration (i.e., principal component analysis (PCA) and its modified algorithms [7]). Comparing between these two methods, feature regeneration method has advantages to more effectively eliminate or reduce redundancy or correlation in the final optimal image feature vector. However, most of medical image data or features have very complicated or heterogeneous distribution patterns, which may not meet the precondition of optimally applying PCA-type feature regeneration methods.

In order to better address this challenge and more reliably regenerate image feature vector for developing CAD schemes of medical images, we in this study investigate and test another feature regeneration method namely, a random projection algorithm. The random projection algorithm is an efficient way to map features into a space with a lower-dimensional subspace, while preserving the distances between points under better contrast. This mapping process is done with a random projection matrix. In the lower space since the distance is preserved, it will be much easier and reliably to classify between two feature classes. Because of its advantages and high performance, random projection algorithms have been tested and implemented in a wide range of engineering applications including handwrite recognition [8], face recognition and detection [9], visual object tracking and recognition [10, 11], and car detection [12]. Thus, motivated by the success of applying random projection algorithms to the complex and nonlinear feature data used in many engineering application domains, we hypothesize that the random projection algorithm can also have advantages when applying to medical images with the heterogeneous feature distributions. To test our hypothesis, we conduct this study to investigate feasibility and potential advantages of applying random projection algorithm to build optimal feature vector and train machine learning model implemented in a new computer-aided diagnosis (CAD) scheme to classify between malignant and benign breast lesions depicting on digital mammograms. For this purpose, a large and diverse image dataset with 1,487 cases is retrospectively assembled and used in this study. From each identified region of interest surrounding a suspicious lesion from the image, an initial feature pool including 181 statistics and texture features is created. Then, machine learning models based on the support vector machines (SVM) are built. To build SVM models, four feature dimensionality reduction methods including random projection algorithm are embedded to the SVM model training and validation process. Finally, lesion classification performance indices are evaluated and compared among the SVMs embedded with 4 different feature vector regeneration methods. The details of the assembled image dataset, the experimental methods of feature regeneration and SVM model optimization, data analysis and performance evaluation results are presented in the following sections.

TABLE I
Case number and percentage distribution of patients age and mammographic density rated by radiologists using BIRADS guidelines.

| | Subgroup | Malignant Cases | Benign Cases |
|---|---|---|---|
| Density BIRADS | 1 | 25 (3.9%) | 58 (6.9%) |
| | 2 | 186 (28.8%) | 262 (31.1%) |
| | 3 | 401 (62.3%) | 502 (59.5%) |
| $p$-value = 0.576 | 4 | 32 (5.0%) | 21 (2.5%) |
| Age of Patients (years old) | A < 40 | 11 (3.4%) | 71 (8.4%) |
| | 40 ≤ A < 50 | 109 (19.2%) | 158 (18.7%) |
| | 50 ≤ A < 60 | 167 (25.6%) | 285 (33.8%) |
| | 60 ≤ A < 70 | 180 (24.4%) | 192 (22.8%) |
| | 70 ≤ A | 177 (27.4%) | 137 (16.3%) |

## II. MATERIALS AND METHODS

### A. Image Dataset

A fully anonymized dataset of full-field digital mammography (FFDM) images acquired from 1,487 patients are retrospectively assembled and used in this study. These patients with the age range from 35 to 80 years old underwent regular annual mammography screening using Selenia Dimensions digital mammography machine (Hologic company). In the original mammogram reading and interpretation, radiologists detected suspicious lesions (soft-tissue masses) in each of these cases and annotated the position of each lesion. All the suspicious lesions had biopsy. From the histopathology examinations of the biopsy-extracted lesion specimens, lesions in 644 cases were confirmed to be malignant, while lesions in 843 cases were benign. Additionally, in this dataset, the majority of cases have two craniocaudal (CC) and mediolateral oblique (MLO) view mammographic images of either left or right breast in which the suspicious lesions are detected by the radiologists, while small fraction of cases just have one CC or MLO image in which the lesions were detected. Overall, 1,197 images depicting malignant lesions and 1,302 images depicting benign lesions are available in this image dataset. Table I summarizes and compares case distribution information of patients' age and mammographic density rated by radiologists using BIRADS guidelines. As shown in the table, patients in benign group are moderately younger than the patients in the malignant group. In BIRADS-rated breast tissue density distribution, there is not a significant difference between the two groups of patients ($p$ = 0.576).

### B. Initial Image Feature Pool with a High Dimensionality

In developing CAD schemes to classify between malignant and benign breast lesions, many different approaches have been investigated and applied to compute image features including those computed from the segmented lesions [13], the fixed regions of interest (ROIs) [14] and the entire breast area [15]. Each approach has advantages and disadvantages. However, their classification performance may be quite comparable with an appropriate training and optimization process. Thus, since this study focus on investigating the feasibility and potential advantages of a new feature dimensionality reduction method namely, the random projection algorithm, we will use a simple approach to compute the initial image features. Specifically, we place a square block (or ROI) of size 150×150 pixels around a suspicious lesion. The ROI is big enough to cover the soft-tissue mass regions included in our large and diverse image dataset.

Since classification between malignant and benign lesions is a difficult task, which depends on optimal fusion of many image features related to tissue density heterogeneity, speculation of lesion boundary, as well as variation of surrounding tissues. Previous studies have demonstrated that statistics and texture features can be used to model these valuable image features including intensity, energy, uniformity, entropy, and statistical moments, etc. Thus, like most CAD schemes using the ROIs with a fixed size as classification targets (including the schemes using deep learning approaches [16]), this CAD scheme also focuses on using the statistics and texture-based image features computed from the defined ROIs and the segmented lesion regions. For this purpose, following methods are used to compute image features that are included in the initial feature pool.

First, from a ROI of an input image, gray level difference method (GLDM) is used to compute the occurrence of the absolute difference between pairs of gray levels divided in a particularly defined distance in several directions. It is a practical way for modeling analytical texture features. The output of this function is four different probability distributions. For an image $I(m,n)$, we consider displacement in different directions like $\delta(d_x, d_y)$, then $\hat{I}(m,n) = |I(m,n) - I(m+d_x, n+d_y)|$ estimates the absolute difference between gray levels, where $d_x, d_y$ are integer values. Now it is possible to determine an estimated probability density function for $\hat{I}(m,n)$ like $f(.|\delta)$ in which $f(i|\delta) = P(\hat{I}(m,n) = i)$. It means for an image with $L$ gray levels, the probability density function is $L$-dimensional. The components in each index of the function show the probability of $\hat{I}(m,n)$ with the same value of the index. In the proposed method implemented in this CAD study, we consider $d_x = d_y = 11$, which is calculated heuristically [17]. The probability functions are computed in four directions ($\varphi = 0, \pi/4, \pi/2, 3\pi/4$), which signifies that four probability functions are computed providing absolute differences in four primary directions. Each of which is used for feature extraction.

Second, a gray-level co-occurrence matrix (GLCM) estimates the second-order joint conditional probability density function. The GLCM carries information about the locations of pixels having similar gray level values, as well as the distance and angular spatial correlation over an image sub-region. To establish the occurrence probability of pixels with the gray level of $i,j$ over an image along a given distance of $d$ and a specific orientation of $\varphi$, we have $P(i,j,d,\varphi)$. In this way, the output matrix has a dimension of the gray levels ($L$) of the image [18]. Like GLDM, we compute four co-occurrence matrices in four cardinal directions ($\varphi = 0, \pi/4, \pi/2, 3\pi/4$). GLCM is rotation invariant. We combine the results of different angles in a summation mode to obtain the following probability density function for feature extraction, which is also normalized to reduce image dependence.

$$P(i,j) = \sum_{\varphi=0,\pi/4,\pi/2,3\pi/4} P(i,j,d=2,\varphi)$$
$$P(i,j) = \frac{P(i,j)}{\sum_i \sum_j P(i,j)}; i,j = 1,2,3,\dots,L \quad (1)$$

Third, a gray level run length matrix (GLRLM) is another popular way to extract textural features. In each local area depicting suspicious breast lesion, a set of pixel values are searched within a predefined interval of the gray levels in several directions. They are defined as gray level runs. GLRM calculates the length of gray-level runs. The length of the run is the number of pixels within the run. In the ROI, spatial variation of the pixel values for benign and malignant lesions may be different, and gray level run is a proper way to delineate this variation. The output of a GLRM is a matrix with elements that express the number of runs in a particular gray level interval with a distinct length. Depending on the orientation of the run, different matrices can be formed [19]. We in this study consider four different directions ($\varphi = 0, \pi/4, \pi/2, 3\pi/4$) for GLRM calculations. Then, just like GLCM, GLRM is also rotation invariant. Thus, the output matrices of different angles in a summation mode are merged to generate one matrix.

Fourth, in addition to the computing texture features from the ROI of the original image in the spatial domain, we also explore and conduct multiresolution analysis, which is a reliable way to make it possible to perform zooming concept through a wide range of sub-bands in more details [20]. Hence, textural features extracted from the multiresolution sub-bands manifest the difference in texture more clearly. Specifically, a wavelet transform is performed to extract image texture features. Wavelet decomposes an image into the sub-bands made with high-pass and low-pass filters in horizontal and vertical directions followed by a down-sampling process. While down-sampling is suitable for noise cancelation and data compression, high-pass filters are beneficial to focus on edge, variations, and the deviation, which can show and quantify texture difference between benign and malignant lesions. For this purpose, we apply 2D Daubechies (Db4) wavelet on each ROI to get approximate and detailed coefficients. From the computed wavelet maps, a wide range of texture features is extracted from principal components of this domain.

Moreover, analyzing geometry and boundary of the breast lesions and the neighboring area is another way to distinguish benign and malignant lesions. In general, benign lesions are typically round, smooth, convex shaped, with well-circumscribed boundary, while malignant lesions tend to be much blurry, irregular, rough, with non-convex shapes [21]. Hence, we also extract and compute a group of features that represent geometry and shape of lesion boundary contour.

Then, we add all computed features as described above to create the initial pool of image features.

*C. Applying Random Projection Method to Generate Optimal Feature Vector*

Before using random projection algorithm to generate an optimal feature vector from the initial image feature pool, we first normalize each feature to make its value distribution between [0, 1] to reduce case-based dependency and weight all features equally. Thus, for each case, we have a feature vector of size $d$, which is valuable to determine that case based on the extracted features as a point in a $d$ dimensional space. For two points like $X = (x_1, \ldots, x_d)$, and $Y = (y_1, \ldots y_d)$, the distance in $d$ dimensional spaces define as:

$$|X - Y| = \sqrt{\sum_{j=1}^{d}(x_j - y_j)^2} \quad (2)$$

In addition, it is also possible to define the volume $V$ of a sphere in a $d$ dimensional space as a function of its radius ($r$) and the dimension of the space as (3). This equation is proved in [22].

$$V(d) = \frac{r^d \pi^{\frac{d}{2}}}{\frac{1}{2}\Gamma(\frac{d}{2})} \quad (3)$$

The matrix of features is normalized between [0, 1]. It means a sphere with $r = 1$ can encompass all the data. An interesting fact about a unit-radius sphere is that as equation (4) shows, as the dimension increase, the volume goes to zero. At the same time, the maximum possible distance between two points stays at 2.

$$\lim_{d \to \infty} \left( \frac{\pi^{\frac{d}{2}}}{\frac{1}{2}\Gamma(\frac{d}{2})} \right) \cong 0 \quad (4)$$

Moreover, based on the heavy-tailed distribution theorem, for a case like $X = (x_1, \ldots, x_d)$ in the space of features, suppose with an acceptable approximation features are independent, or nearly perpendicular variables as mapped to different axes, with $E(x_i) = p_i$, $\sum_{i=1}^{d} p_i = \mu$ and $E|(x_i - p_i)^k| \leq p_i$ for $k = 2, 3, \ldots, \lfloor t^2/6\mu \rfloor$, then, it is possible to prove that:

$$prob(\left|\sum_{i=1}^{d} x_i - \mu\right| \geq t) \leq Max\left(3e^{\frac{-t^2}{12\mu}}, 4 \times 2^{\frac{-t}{e}}\right) \quad (5)$$

We can perceive that the farther the value of $t$ increases, the smaller the chance of having a point out of that distance, which means that $X$ would be concentrated around the mean value. Overall, based on equations (4), and (5) with an acceptable approximation, all data are encompassed in a sphere of size one, and they are concentrated around their mean value. As a result, if the dimensionality is high, the volume of the sphere is close to zero. Hence, the contrast between the cases is not enough for a proper classification.

Above analysis also indicates the more features included in the initial feature vector, the higher the dimension of the space is, and the more data is concentrated around the center, which makes it more difficult to have enough contrast between the features. A powerful technique to reduce the dimensionality while approximately preserves the distance between the points, which implies approximate preservation of the highest amount of information, is the key point that we are looking for. If we adopt a typical feature selection method and randomly select a k-dimensional sup-space of the initial feature vector, it is possible to prove that all the projected distances in the new space are within a determined scale-factor of the initial d-dimensional space [23]. Hence, although some redundant features are removed, the final accuracy may not increase, since contrast between the points may still be not enough to present a robust model.

To address this issue, we take advantage of Johnson-Lindenstrauss Lemma to optimize the feature space. Based on the idea of this lemma, for any $0 < \epsilon < 1$, and any number of cases as $N$, which are like the points in $d$-dimensional space ($R^d$), if we assume $k$ as a positive integer, it can be computed as:

$$k \geq 4 \frac{\ln N}{(\frac{\epsilon^2}{2} - \frac{\epsilon^3}{3})} \quad (6)$$

Then, for any set $V$ of $N$ points in $R^d$, for all $u, v \in V$, it is possible to prove that there is a map, or random projection function like $f: R^d \to R^k$, which preserves the distance in the following approximation [24]:

$$(1 - \epsilon)|u - v|^2 \leq |f(u) - f(v)|^2 \leq (1 + \epsilon)|u - v|^2 \quad (7)$$

Another arrangement of this formula is like:

$$\frac{|f(u) - f(v)|^2}{(1 + \epsilon)} \leq |u - v|^2 \leq \frac{|f(u) - f(v)|^2}{(1 - \epsilon)} \quad (8)$$

As these formulas show the distance between the set of points in the lower-dimension space is approximately close to the distance in high-dimensional space. This Lemma states that it is possible to project a set of points from a high-dimensional space into a lower dimensional space, while the distances between the points are nearly preserved.

It implies that if we project the initial group of features into a space with a lower-dimensional subspace using the random projection method, the distances between points are preserved under better contrast. This may help better classify between two feature classes representing benign and malignant lesions with low risk of overfitting. In this study, we investigate and demonstrate whether using this random projection algorithm can yield better result as comparable to other feature dimensionality reduction approaches (i.e., the popular principal component analysis).

*D. Experiment of Feature Combination and Dimensionality Reduction*

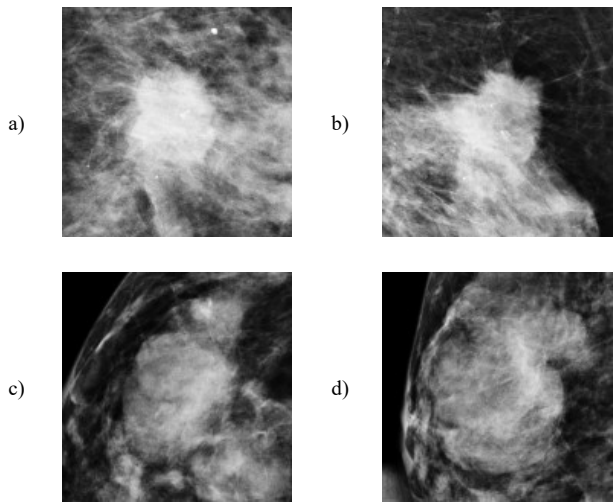

Figure1. Example of 4 extracted ROIs with the detected suspicious soft-tissue masses (lesions) in ROI center. a,b) 2 ROIs involving malignant lesions and c,d) 2 ROIs involving benign lesions.

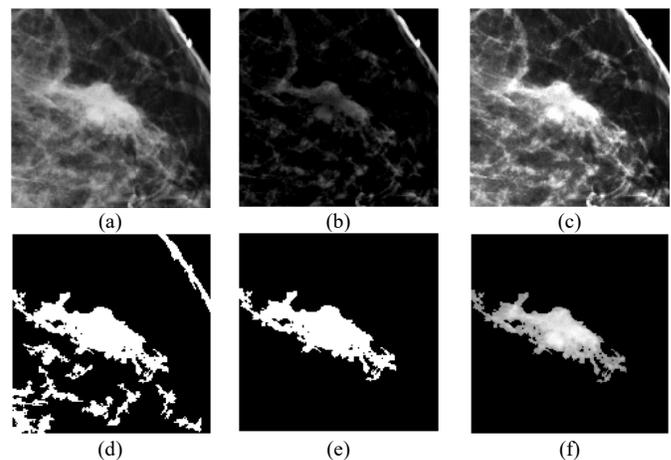

Figure 2. Example to illustrate lesion segmentation, which include a) the original ROI, b) absolute difference of ROI from low-pass filtered version, c) combination of (a) and (b) which gives the suspicious regions better contrast to the background, d) output of morphological filtering, e) blob with the largest size is selected (a binary version of the lesion), and f) finally segmented lesion area. It is output of mapping (e) to (a).

First, the proposed CAD scheme applies an image preprocessing step to the whole images in the dataset to read them one by one, and based on the annotated location of the lesions, cut the ROI area as a square of size 150×150 in which the centers of the lesion and ROI overlap. In our study, a heuristic method is applied to select ROI size. Basically, the different ROI sizes (i.e., from 128×128 to 180×180 pixel range) are examined and compared. From the experiments, we observe that the ROIs with size of 150×150 has the best classification results applying to this large and diverse dataset, which reveals that this is the most efficient size. Figure 1 shows examples of two malignant lesion regions and two benign lesion regions. After ROI determination, all the images in the dataset are saved in Portable Network Graphics (PNG) format with 16 bits in the lossless mode for the feature extraction phase.

Next, the CAD scheme is applied to segment lesion from the background. For this process, CAD first defines a low pass filter with a window-size of 30 and utilizes it to the whole ROI. The absolute difference of ROI from the filtered version of the image is calculated, which is an image with no background. If mapping the segmented region back to the original ROI, lesion and the other suspicious regions are highlighted with higher contrast. After that, applying morphological filters (i.e., opening and closing), different blobs are detected in the ROI area. The blob with the largest size is selected as suspicious lesion. Figure 2 shows an example of applying this algorithm to locate and segment suspicious lesion from the surrounding tissue background.

Then, CAD scheme is applied to extract and compute several sets of the relevant image features from the entire ROI and the segmented lesion as well. The first group of features are the pixel value (or density) related statistics features as summarized in Table II. These 20 statistics features are repeatedly computed from three types of images namely, 1) the entire ROI of the original images (as shown in figure 2(a)), 2) the segmented lesion region (as shown in figure 2(f)), and 3) all segmented blobs with pixel numbers greater than 50 (as shown in figure. 2(d)). Thus, this group of features includes 60 statistics features.

The second group of features is computed from the GLRLM matrix of the ROI area. For this purpose, 16 different quantization levels are considered to calculate all probability functions in four different directions from the histograms. After combining the probability functions, on rotation invariance version of them, the following group of features is computed. Features are short-run emphasis, long-run emphasis, gray level non-uniformity, run percentage, run-length non-uniformity, low gray level run emphasis, and high gray level run emphasis. Hence, this group of features includes seven GLRM-based features.

The third group of features includes GLDM based features computed from the entire ROI. Specifically, we select a distance value of 11 pixels for the inter-sample distance calculation. CAD computes four different probability density functions (PDFs) based on the image histogram calculation in different directions. The PDF ($p$) with ($\mu$) as the mean of the population, standard deviation, root mean square level, and the first four statistical moments ($n = 1, 2, 3, 4$) with the following equation are calculated as features.

$$\hat{m}_n = \sum_{i=1}^{N} p_i (x_i - \mu)^n \qquad (9)$$

It is an unbiased estimate of $n^{th}$ moment possible to

*Table II*
*List of the computed Features on ROI Area*

| Feature category | Feature Description |
|---|---|
| Features computed On the Whole ROI | 1.Mean, 2. variance, 3. skewness, 4. kurtosis, 5. entropy, 6. correlation, 7. energy, 8. root mean square level, 9. uniformity, 10. max, 11. min, 12. median, 13. range, 14. mean absolute deviation, 15. Contrast, 16. homogeneity, 17. smoothness, 18. inverse difference movement, 19. suspicious regions volume, 20. standard deviation. |

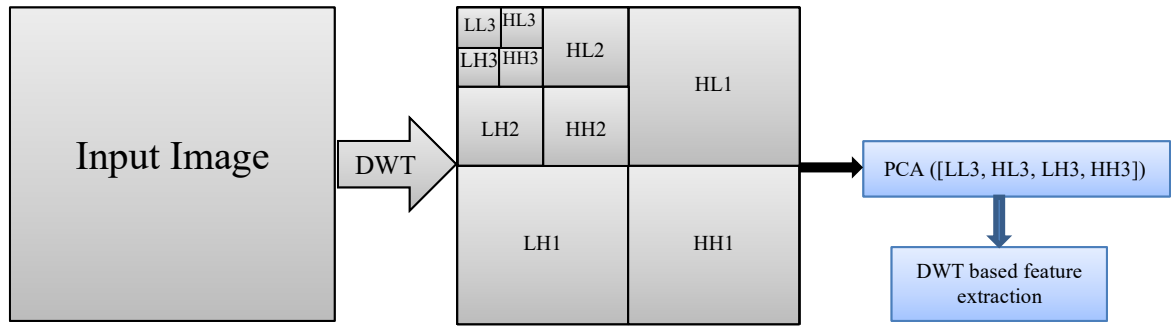

Figure 3. Wavelet based feature extraction. Wavelet decomposition is applied three times to make the images compress as possible. Then PCA is adopted as another way of data compression.

calculate by:

$$m_n = \int_{-\infty}^{\infty} p(x)x^n dx \quad (10)$$

As shown in equation 10, $p(x)$ is weighted by $x^n$. Hence, any change in the $p(x)$ is polynomially reinforced in the statistical moments. Thus, any difference in the four PDFs computed from malignant lesions is likely to be polynomially reinforced in the statistical moments of the computed coefficients. Six features from each of four GLDM based PDFs make this feature group, which has total 24 features.

The fourth group of features computes GLCM based texture feature. Based on the method proposed in the previous study [25], our CAD scheme generates a matrix of 44 textural features computed from GLCM matrix based on all GLCM based equations proposed in [18]. In this way any properties of the GLCM matrix proper for the classification purpose is granted. Hence, this group contains 44 features computed from the entire ROI.

The fifth group of features includes wavelet-based features. The Daubechies wavelet decomposition is accomplished on the original ROI (i.e., figure 2(a)). Figure 3 shows a block diagram of the wavelet-based feature extraction procedure. The last four sub-bands of wavelet transform are used to build a matrix of four sub-bands in which principal components of this matrix are driven for feature extraction and computation. The computed features are listed in table III. We also repeat the same process to compute wavelet-based feature from the segmented lesion (i.e., figure 2(f)). As a result, this feature group includes 23 wavelet-based image features.

Last, to address the differences between morphological and structural characteristics of benign and malignant lesions, another group of geometrical based features is derived and computed from the segmented lesion region. For this purpose, a binary version of the lesion, like what we showed in figure 2 (e), is first segmented from the ROI area. Then, all the properties listed in table IV are calculated from the segmented lesion region in the image using the equations reported in [26].

By combining all features computed in above 6 groups, CAD scheme creates an initial pool of 181 image features. Then, a random projection algorithm is applied to reduce feature dimensionality and generate an optimal feature vector.

Table III
List of Wavelet-based Features

| Feature category | Feature Description |
|---|---|
| Features computed On the Whole ROI | 1. Contrast, 2. Correlation, 3. Energy, 4. Homogeneity, 5. Mean, 6. Standard deviation, 7. Entropy, 8. Root mean square level, 9. Variance, 10. Smoothness, 11. Kurtosis, 12. Skewness, 13. IDM |

Table IV
List of Geometrical Features

| Feature category | Feature Description |
|---|---|
| Features computed On the Whole ROI | 1. Area, 2. Major Axis Length, 3. Minor Axis Length, 4. Eccentricity, 5. Orientation, 6. Convex Area, 7. Circularity, 8. Filled Area, 9. Euler Number, 10. Equivalent Diameter, 11. Solidity, 12. Extent, 13. Perimeter, 14. Perimeter Old,15. Max Feret Diameter,16. Max Feret Angle,18. Min Feret Diameter,19. Min Feret Angle, 20. Roundness Ratio |

For this purpose, we utilize sparse random matrix as the projection function to achieve the criteria as defined in equation (7). Sparse random matrix is a memory efficient and fast computing way of projecting data, which guarantees the embedding quality of this idea. To do so, if we define $s = 1/density$, the components of the matrix as random matrix elements (RME) are:

$$RME = \begin{cases} -\sqrt{\frac{s}{n_{components}}}, & 1/2s \\ 0, & \text{with probability} \quad 1 - 1/s \\ \sqrt{\frac{s}{n_{components}}}, & 1/2s \end{cases} \quad (11)$$

In this process, we select $n_{components}$, which is the size of the projected subspace. As recommended in [27], we consider number of non-zero elements to the minimum density.

E. *Development and Evaluation of Machine Learning Model*

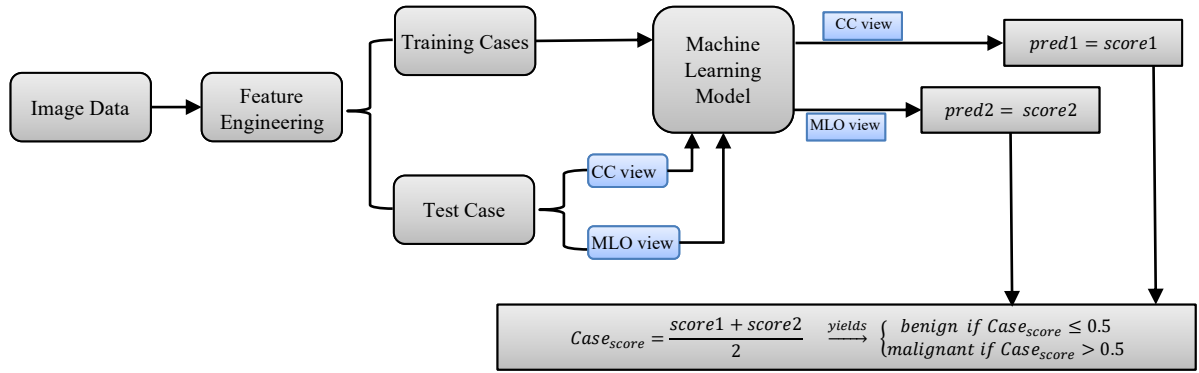

Figure 4. Overall classification flow

After processing images and computing image features from all 1,197 ROIs depicting malignant lesions and 1,302 ROIs depicting benign lesions, we build machine learning model to classify between malignant and benign lesions by taking following steps or measures. Figure 4 shows a block diagram of the machine learning model along with the training and testing process. First, although many machine learning models (i.e., artificial neural networks, K-nearest neighborhood network, Bayesian belief network, support vector machine and others) have been investigated and used to develop CAD schemes, based on our previous research experience [15], we adopt the support vector machine (SVM) to train a multi-feature fusion based machine leaning model to predict the likelihood of lesions being malignancy in this study. Under a grid search and hyperparameter analyses, linear kernel implemented in SVM model can also achieve a low computational cost and high robustness in prediction results as well.

Second, we apply the random projection algorithm to reduce the dimensionality of the image feature space and map them to the most efficient feature vector as input features of the SVM model. To demonstrate the potential advantages of using the random projection algorithm in developing machine learning models, we build and compare 5 SVM models, which using all 181 image features included in the initial feature pool, and embedding 4 feature dimensionality reduction methods including (1) random projection algorithm (RPA), (2) principle component analyses (PCA), (3) nonnegative matrix factorization (NMF), and (4) Chi-squared (Chi2).

Third, to increase size and diversity of training cases, as well as reduce the potential bias in case partitions, we use a leave-one-case-out (LOCO) based cross-validation method to train SVM model and evaluate its performance. The feature dimensionality reduction method as discussed in the second step is also embedded in this LOCO iteration process to train the SVM. This can diminish the potential bias in the process of feature dimensionality reduction and machine learning model training as we demonstrated in our previous study [28]. When the random projection algorithm is embedded in the LOCO based model training process, it helps generate a feature vector independent of the test case. Thus, the test case is unknown to both random projection algorithm and SVM model training process. In this way, in each LOCO iteration cycle, the trained SVM model is tested on a truly independent test case by generating an unbiased classification score for the test case. As a result, all SVM-generated classification scores are independent of the training data.

Fourth, since majority of lesions detected in two ROIs from CC and MLO view mammograms, in the LOCO process, two ROIs representing the same lesion will be grouped together to be used for either training or validation to avoid potential bias. After training, ROIs in one remaining case will be used to test the machine learning model that generates a classification score to indicate the likelihood of each testing ROI depicting a malignant lesion. The score ranges from 0 to 1. The higher score indicates a higher risk of being malignant. In addition to the classification score of each ROI, a case-based likelihood score is also generated by fusion of two scores of two ROIs representing the same lesion depicting on CC and MLO view mammograms.

Fifth, a receiver operating characteristic (ROC) method is applied in the data analysis. Area under ROC curve (AUC) is computed from the ROC curve and utilized as an evaluation index to evaluate and compare performance of each SVM model to classify between the malignant and benign lesions. Then, we also apply an operating threshold of T = 0.5 on the SVM-generated classification scores to classify or divide all testing cases into two classes of malignant and benign cases. By comparing to the available ground-truth, a confusion matrix for the classification results is determined for each SVM. From the confusion matrix, we compute classification accuracy, sensitivity, specificity, and odds ratio (OR) of each SVM model based on both lesion region and case. In the region-based performance evaluation, all lesion region are considered independent, while in the case-based performance evaluation, the average classification score of two matched lesion regions (if the lesions are detected and marked by radiologists in both CC and MLO view) is computed and used. In this study, all pre-processing and feature extraction steps to make the matrix of features are conducted using MATLAB R2019a package.

## III. RESULTS

Figure 5 shows a malignant case as an example in which the lesion center is annotated by radiologists in both CC and MLO view mammograms. Based on the marked center, we plot two square areas on two images in which image features are computed by the CAD scheme. Using the whole feature vector of 181 image features, the SVM-model generates the following classification scores to predict the likelihood of two lesion regions on two view images being malignant, which are $S_{CCview} = 0.685$, and $S_{MLOview} = 0.291$. The case-based classification score is $S_{Case} = 0.488$. When using the feature vectors generated by the random projection algorithm, the SVM-model generates two new classification scores of these two lesion regions, which are $S_{CCview} = 0.817$, and $S_{MLOview} = 0.375$. Thus, the case-based classification score is $S_{Case} = 0.596$. As a result, using the SVM model trained using all 181 image features misclassifies this malignant lesion into benign when an operating threshold (T = 0.5) is applied, while the SVM model trained using the embedded random projection algorithm increases the classification scores for both lesion regions depicting on CC and MLO view images. As a result, it is correctly classified as malignant with the case-based classification score greater than the operating threshold.

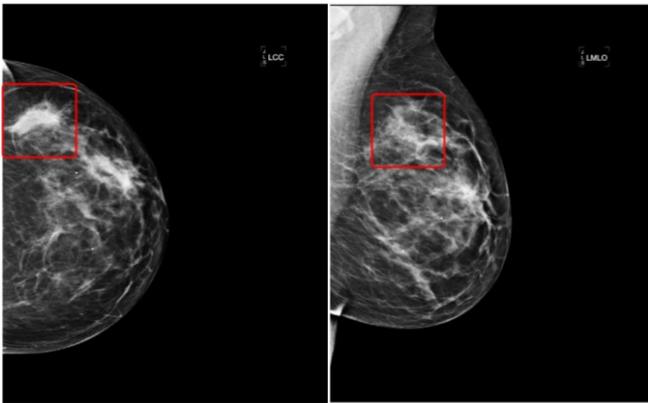

Figure 5. A malignant case annotated by radiologists in both CC and MLO views. The annotated mass is squared in each view.

TABLE V
Summary of average number of image features used in 5 different SVM models and classification performance (AUC) based on both region and case-based lesion classification.

| Feature sub-groups | Number of features | AUC |
|---|---|---|
| Original features, region based | 181 | 0.72 |
| Original features, case based | 181 | 0.74 |
| NMF, region based | 100 | 0.73 |
| NMF, case based | 100 | 0.77 |
| Chi2, region based | 76 | 0.73 |
| Chi2, case based | 76 | 0.75 |
| PCA, region based | 83 | 0.75 |
| PCA, case based | 83 | 0.79 |
| RP, region based | 80 | 0.78 |
| RP, case based | 80 | 0.84 |

Table V shows and compares the average number of input features used to train 5 SVM models with and without embedding different feature dimensionality reduction methods, lesion region-based and case-based classification performance of AUC values. When embedding a feature dimensionality reduction algorithm, the size of feature vectors in different LOCO-based SVM model training and validation cycle may vary. Table V shows that average number of features are reduced from original 181 features to 100 or less. When using random projection algorithm, the average number of features is 80. From both table V and figure 6, which shows and compares the corresponding ROC curves, we can observe that SVM models trained using an embedded feature dimensionality reduction method produces the higher or improved classification performance as comparing to the SVM model trained using the initial feature pool of 181 features. Among them, the SVM model embedding with a random projection algorithm achieves a significantly higher performance with a case-based AUC value of 0.84±0.01 than other SVM models without using feature dimensionality reduction and embedded with other three feature dimensionality reduction methods namely, principle component analyses (PCA), nonnegative matrix factorization (NMF) and Chi-squared (Chi2) ($p < 0.05$). In addition, the data in table V and ROC curves in figure 6 also indicate that lesion case-based classification yields higher performance than region-based classification performance, which indicates that using and combining image features from two-view mammograms is helpful.

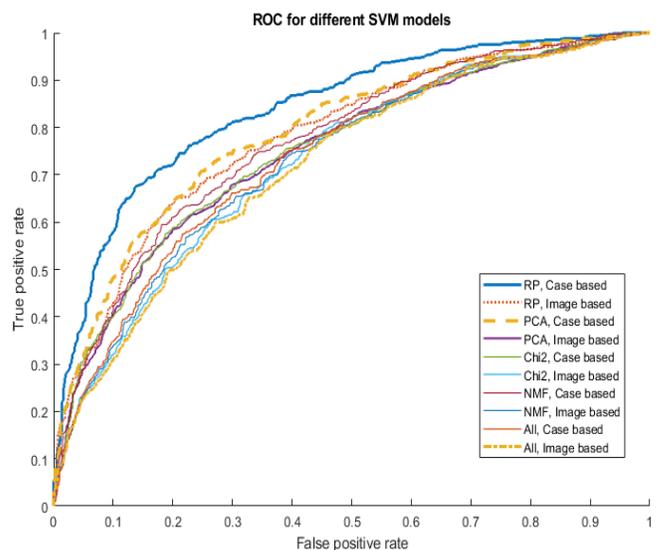

Figure 6. Comparison of 10 ROC curves generated using 5 SVM models and 2 scoring (region and case-based) methods to classify between malignant and benign lesion regions or cases.

Table VI presents 5 confusion matrices of lesion case-based classification using 5 SVM-models after applying the operating threshold (T = 0.5). Based on this table, several lesion classification performance indices like sensitivity, specificity, and odds ratio are measured and shown in table VII. This table also shows that the SVM model trained based on the feature vector generated by the random projection algorithm yields the highest classification accuracy comparing to the other 4 SVM models trained using feature vectors

TABLE VI

Five Confusion matrices of case-based lesion classification using 5 different SVM models to classify between benign and malignant cases.

| Feature Group | Predicted | Actual Positive | Actual Negative |
|---|---|---|---|
| Original features | Positive | 399 | 212 |
|  | Negative | 245 | 631 |
| NMF | Positive | 406 | 173 |
|  | Negative | 238 | 670 |
| Chi2 | Positive | 405 | 194 |
|  | Negative | 239 | 649 |
| PCA | Positive | 436 | 197 |
|  | Negative | 208 | 646 |
| RPA | Positive | 452 | 177 |
|  | Negative | 192 | 666 |

TABLE VII

Summary of the lesion case-based classification accuracy, sensitivity, specificity, and odd ratio of using 5 SVMs trained using different groups of optimized features.

| Feature sub-group | Accuracy (%) | Sensitivity (%) | Specificity (%) | Odds Ratio |
|---|---|---|---|---|
| Original features | 69.3 | 62.0 | 75.0 | 4.85 |
| NMF | 72.4 | 63.1 | 79.5 | 6.61 |
| Chi2 | 70.9 | 63.0 | 77.1 | 5.67 |
| PCA | 72.8 | 68.0 | 76.6 | 6.87 |
| RPA | **75.2** | **70.2** | 79.0 | **8.86** |

generated either based on other three feature dimensionality reduction methods or the original feature pool of 181 features.

## IV. DISCUSSION

Mammography is a popular imaging modality used in breast cancer screening. However, due to the heterogeneity of breast lesions and dense fibro-glandular tissue, it is difficult for radiologists to accurately predict or determine the likelihood of the suspicious lesions being malignant. As a result, mammography screening has very high false-positive recall rates and majority of biopsies are approved to be benign [29]. Thus, to help increase specificity of breast lesion classification and reduce the unnecessary biopsies, developing CAD schemes to assist radiologists more accurately and consistently classifying between malignant and benign breast lesions remains an active research topic that continues attracting broad research interest in medical imaging informatics or CAD field [30]. In this study, we develop and assess a new CAD scheme of mammograms to predict the likelihood of the detected suspicious breast lesions being malignant. This study has following unique characteristics as comparing to other previous CAD studies reported in the literature.

First, previous CAD schemes of mammograms computed image features from either the segmented lesion regions or the regions with a fixed size (i.e., squared ROIs to cover lesions with varying sizes). Both approaches have advantages and disadvantages. Due to the difficulty to accurately segment subtle lesions with fuzzy boundary, the image features computed from the automatically segmented lesions may not be accurate or reproducible, which reduces the accuracy of the computed image features to represent actual lesion regions. When using the fixed ROIs (including the new deep learning based CAD schemes [16, 31]), although it can avoid the potential error in lesion segmentation, it may lose and reduce the weight of the image features that are more relevant to the lesions due to the potential heavy influence of irregular fibro-glandular tissue distribution surrounding the lesions with varying sizes. In this study, we tested a new approach that combines image features computed from both a fixed ROI and the segmented lesion region. In addition, comparing to the most of previous CAD studies as surveyed in the previous study, which used several hundreds of malignant and benign lesion regions [32], we assemble a much larger image dataset with 1,847 cases or 2,499 lesion region (including 1,197 malignant lesion regions and 1,302 benign lesion regions). Despite using a much larger image dataset, this new CAD scheme yields a higher classification performance (AUC = 0.84±0.01) as comparing to AUC of 0.78 to 0.82 reported in our previous CAD studies that using much smaller image dataset (<500 malignant and benign ROIs or images) [16, 33]. Thus, although it may be difficult to directly compare performance of CAD schemes tested using different image datasets as surveyed in [32], we believe that our new approach to combine image features computed from both a fixed ROI and the segmented lesion region has advantages to partially compensate the potential lesion segmentation error and misrepresentation of the lesions related image features, and enable to achieve an improved or very comparable classification performance.

Second, since identifying a small, but effective and non-redundant image feature vector plays an important role in CAD development to train machine learning classifiers or models, many feature selection or dimensionality reduction methods have been investigated and applied in previous studies. Although these methods can exclude many redundant and low-performed or irrelevant features in the initial pool of features, the challenge of how to build a small feature vector with orthogonal feature components to represent the complex and non-linear image feature space remains. For the first time, we in this study introduce the random projection algorithm to the medical imaging informatics field to develop CAD scheme. Random projection is a technique that maximally preserves the distance between the sub-set of points in the lower-dimension space. As explained in the Introduction section, in the lower space under preserving the distance between points, classification is much more robust with low risk of overfitting. This is not only approved by the simulation or application results reported in previous studies, it is also confirmed in this study. The results in table V show that by using the optimal feature vectors generated by random projection algorithm, the SVM model yields significantly higher classification performance in comparison with other SVM models trained using either all initial features or other feature vectors generated by other three popular feature selection and dimensionality reduction methods. Using random projection algorithm boosts the AUC value from 0.72 to 0.78 in comparison with the original feature vector in the lesion region-based analysis, and from 0.74 to 0.84 in the lesion case-base evaluation, which also enhances the classification accuracy from 69.3% to 75.2%, and approximately doubling the odds ratio from 4.85 to 8.86 (table VI). Thus, the study results confirm that random projection algorithm is a very promising technique applicable to generate optimal feature vectors for training machine learning models used in CAD of medical images.

Third, since the heterogeneity of breast lesions and surround fibro-glandular tissues distributed in 3D volumetric space, the segmented lesion shape and computed image features often vary significantly in two projection images (CC and MLO view), we investigate and evaluate CAD performance based on single lesion regions and the combined lesion cases if two images of CC and MLO views were available and the lesions are detectable on two view images. Table V shows and compares lesion region-based and case-based classification performance of 5 SVM models. The result data clearly indicate that instead of just selecting one lesion region for likelihood prediction, it would be much more accurate when the scheme processes and examines two lesion regions depicting on both CC and MLO view images. For example, when using the SVM trained with the feature vectors generated by the random projection algorithm, the lesion case-based classification performance increases 7.7% in AUC value from 0.78 to 0.84 as comparing to the region-based performance evaluation.

Last, although the study has tested a new CAD development method and yielded encouraging results to classify between malignant and benign breast lesions, we realize that the reported study results are made on a laboratory-based retrospective image data analysis process with several limitations. First, although the dataset used in this study is relatively large and diverse, whether this dataset can sufficiently represent real clinical environment or breast cancer population is unknown or not tested. Second, in this retrospective study, it has higher ratio between malignant and benign lesions, which may be different from the false-positive recall rates in the clinical practices. Thus, the reported AUC values may also be different from the real clinical practice, which needs to be further tested in future prospective clinical studies. Third, in the initial pool of features, we only extracted a limited number of 181 statistics and textural features, which are much less than the number of features computed based on recently developed radiomics concept and technology [3, 4]. Thus, more texture features can be explored in future studies to increase diversity of the initial feature pool, which may also increase the chance of selecting or generating more optimal features.

## V. Conclusions

In summary, due to the difference between human vision and computer vision, it is difficult to accurately identify a small set of optimal and non-redundant features computed by CAD schemes. To address this issue, two approaches have been attracting broad research interest recently. One uses deep learning models to automatically search for optimal image features and build classification models. The disadvantage of this approach is requirement of very big training and validation image datasets to build robust deep learning models, which are often unavailable in medical image fields. Another approach is use of radiomics concept and method, which generates a large initial feature pool followed by applying a feature selection method to select a small set of features. This is often a suboptimal approach because many features can still be correlated. In this study, we investigate feasibility of applying a new approach based on random projection algorithm aiming to generate optimal feature vectors for training machine learning models implemented in the CAD schemes of mammograms to classify breast lesions. This approach creates orthogonal feature space that can avoid or minimize feature correlation. By comparing with other three popular feature dimensionality reduction methods, the study results demonstrate that using random projection algorithm enables to generate an optimal feature vector to build a machine learning model, which yields significantly higher classification performance. Last, since building an optimal feature vector is an important precondition of building optimal machine learning models, this approach is not only limited to the CAD scheme of mammograms, it can also be adopted and used by researchers to develop and optimize CAD schemes of other medical images to detect and diagnose different types of cancers or diseases in the future